%% file: steprecovery.tex
\documentclass{svproc}
\usepackage[numbers]{natbib}
\usepackage{amssymb}
\usepackage{amsmath}
\usepackage{commath}
\usepackage{mathtools}

\usepackage{bbm}
\usepackage{hyperref}
\usepackage{import}
\usepackage{paralist}
\usepackage[pdf]{svg}
\usepackage{graphics} 
\usepackage{epsfig} 
\usepackage{url}
\usepackage{multicol}
\usepackage{footmisc}
\usepackage{IEEEtrantools}
\IEEEeqnarraydefcolsep{0}{1.5\parindent}

\DeclareMathOperator*{\minimize}{minimize}

\usepackage{subfloat}

\let\oldbibliography\bibliography
\renewcommand{\bibliography}[1]{{%
		\let\chapter\section
		\oldbibliography{#1}}}

\begin{document}
\mainmatter              
\title{A Receding Horizon Push Recovery Strategy\\ for Balancing the iCub Humanoid Robot}
\titlerunning{A Receding Horizon Push Recovery Strategy}  
%
\author{Stefano Dafarra, Francesco Romano \and Francesco Nori}
\authorrunning{Dafarra et al.} 
%
\tocauthor{Stefano Dafarra, Francesco Romano, Francesco Nori}
\institute{iCub Facility Department, Istituto Italiano di Tecnologia, 16163 Genova,
	Italy,\\
\email{name.surname@iit.it}}

\maketitle              

\begin{abstract}
    Balancing and reacting to strong and unexpected pushes is a critical requirement for humanoid robots. 
    We recently designed a capture point based approach which interfaces with a momentum-based torque controller and we implemented and validated it on the iCub humanoid robot.
    In this work we implement a Receding Horizon control, also known as Model Predictive Control, to add the possibility to predict the future evolution of the robot, especially the constraints switching given by the hybrid nature of the system.
    We prove that the proposed MPC extension makes the step-recovery controller more robust and reliable when executing the recovery strategy.
    Experiments in simulation show the results of the proposed approach.
\end{abstract}
\begin{footnotesize}
	This work was supported by the FP7 EU project CoDyCo (No. 600716 ICT 2011.2.1 Cognitive Systems and Robotics) and Horizon 2020 EU project An.Dy. (No. 731540 Research and Innovation Programme)
\end{footnotesize}
%
\import{tex/}{soa}
\import{tex/}{icub}
\import{tex/}{model}
\import{tex/}{constraints}
\import{tex/}{cost_fcn}
\import{tex/}{simulation}
\import{tex/}{conclusions}

%
%
\bibliography{Bibliography}

\end{document}

%% file: tex/soa.tex
\section{Introduction}\label{c6sec:soa}
A humanoid robot is generally built with the final intent of having a machine which can help humans performing boring or dangerous tasks. The adoption of a two legged design ideally allows the robot to share the same environment of humans. On the other hand, embedding relatively simple motor skills on such complicated machines is not a straightforward task.
A recent trend consists in facing this problem through the adoption of optimal control techniques, like Model Predictive Controllers (MPC).
The peculiarity of MPC resides in the possibility of inserting a feedback from the plant into the optimization procedure, through the ``Receding Horizon Principle'' \cite{Mayne90MPC}. 

Changes in the constraints set, e.g. when walking, result in a different evolution of the constrained dynamical system, making the overall system hybrid \cite{lygeros1999hybrid}. 
Model Predictive Controllers are appealing for controlling hybrid systems \cite{lazar2006stabilizing}. In fact, while predicting future states, MPC allow to cope with both time- and state-dependent switching, eventually undertaking actions to prepare for impending dynamics variations. Nevertheless, MPC is affected by the same problematics related to hybrid systems integration, which are indeed open research problems.

In literature the complexity of the robot dynamics is usually boiled down to simplified models like the Linear Inverted Pendulum \cite{Kajita2001}, in order to devise simple strategies. 
In this context, in \cite{wieber2006trajectory} a model predictive controller has been employed to stabilize walking patterns. It is also worth mentioning the Capture Point framework \cite{Pratt2006} which allows to elaborate conditions about  the possibility of the robot to maintain the upright position. In \cite{krause2012stabilization} it has also been applied in conjunction with MPC. 

In our approach we decided to take a different direction, taking into account the dynamic evolution of the center of mass (CoM) rather then adopting simple models. In addition, with respect to \cite{dai2014whole,herzog2015trajectory,daiplanning} we define a reactive planner which directly interfaces with the momentum-based whole body torque controller already implemented on iCub \cite{nava16,Frontiers2015}. The hybrid nature of the system is taken into account by means of time-varying constraints. 

%% file: tex/icub.tex
\section{Background} \label{sec:background}
Throughout this paper we adopt the following notation. 
\begin{itemize}
	\item $\mathcal{I}$ represents an inertial reference frame with the origin placed on the ground, with the $z-$axis pointing against the gravity and the $x-$axis oriented frontally with respect to the robot. 
	\item $x_\text{CoM} \in  \mathbb{R}^3$ is the position of the center of mass with respect to $\mathcal{I}$.
	\item $\norm{x}^2_W = x^\top W x$ is the weighted square norm of $x$.
	\item ${1}_n$ represents a $n \times n$ identity matrix. $0_{n \times m} \in \mathbb{R}^{n\times m}$ is a zero matrix while $0_n = 0_{n \times 1}$ is a zero column vector of size $n$.
	\item $x^\wedge \in \mathbb{R}^{3 \times 3}$ denotes the skew-symmetric matrix such that $x^\wedge y = x \times y$, where $\times$ denotes the cross product operator in $\mathbb{R}^3$. 
\end{itemize}

\subsection{Momentum-based whole-body torque control}
\label{sec:momentum}

The momentum-based balancing controller has a hierarchical structure accomplishing two different control tasks.
The most important task is in charge of controlling the robot linear and angular momentum.  

We denote with $H = [H_\text{lin}^\top, H_\text{ang}^\top]^\top \in \mathbb{R}^6$ the robot linear and angular momentum (computed around the CoM). 
Its rate of change is: 
\begin{equation}\label{eq:momentum}
\dot{H} = \sum_{i = 1}^{n_c} {}^{\text{CoM}}X_i f_i + m \bar{g}
\end{equation}
where $f_i \in \mathbb{R}^6 = [\mathbbm{f}_i^\top, \tau_i\top]^\top$ is the $i$-th of the $n_c$ contact wrenches. They are measured in a frame fixed with the contact surface. Matrices ${}^\text{CoM}X_i \in \mathbb{R}^{6 \times 6}$ transform these wrenches in a frame oriented as the inertial frame $\mathcal{I}$, with the origin located on the center of mass position. Finally the robot total mass is indicated with $m$ while $\bar{g} \in \mathbb{R}^6$ is the 6D gravity acceleration vector.
By controlling the robot momentum through joint torques, the first task enables us to consider $f = [f_1^\top, \cdots, f_{n_c}^\top]^\top$ as virtual control inputs.

The second objective is responsible of tracking joint references, without interfering with the momentum control. In addition, it guarantees the  stability of the zero dynamics resulting from the convergence of the first task. For additional details see the results in \cite{nava16}.

\subsection{Optimal control problem}
A continuous time optimal control problem can be stated as follows:
\begin{IEEEeqnarray}{rClR}
	\IEEEyesnumber \phantomsection\label{c6eq:ocdef}
	\minimize_{x(t),u(t)} &~& \Gamma = \int_{0}^{t_f}L(x(t),u(t))\mathrm{d}t + E(x(t_f)) \IEEEyessubnumber\\
	\text{subject to}  && \dot{x}(t) = \mathrm{f}(x(t),u(t)),\quad x(0) = x_0 \quad &t\in \left[0,t_f\right] \IEEEyessubnumber \label{c6eq:genmodel}\\ 
	&&g(x(t),u(t)) \leq 0, \quad  &t\in \left[0,t_f\right] \IEEEyessubnumber
\end{IEEEeqnarray}
where $t_f$ is the ``control horizon'', $u \in \mathbb{R}^m$ is the control variable, $x \in \mathbb{R}^n$ is the state variable (whose initial value is $x_0$) and $f:\mathbb{R}^n \times \mathbb{R}^m \rightarrow\mathbb{R}^n$ the corresponding state evolution. 
$L(x(t),u(t))$ is the integral cost, while $E(x(t_f))$ weights $x$ at the end of the horizon. 
Finally $g:\mathbb{R}^n \times \mathbb{R}^m\rightarrow\mathbb{R}^{n_g}$ is the constraints function, embracing both equalities and inequalities.

%% file: tex/model.tex
\section{Problem Formulation} \label{sec:problem_formulation}
\subsection{The model} \label{sec:model}
In this work we focus on the dynamics of the center of mass.
If we consider our case of interest, where no contacts are available except the two applied at the feet, we can rewrite Eq. \ref{eq:momentum} as:
\begin{equation}\label{c6eq:model_2f}
\begin{bmatrix}
\ddot{x}_\text{CoM}\\
\dot{H}_\text{ang}
\end{bmatrix} {=} \begin{bmatrix}
				m^{-1}{1}_3 & 0_{3\times3}\\
				0_{3\times3} & {1}_3
				\end{bmatrix}\left[
				{}^{\text{CoM}}X_{l} \: {}^{\text{CoM}}X_{r}
				\right]\begin{bmatrix}
									f_l\\f_r
							\end{bmatrix} {+} \bar{g}
\end{equation}
where the indexes $l$ and $r$ are relative to the left and right foot respectively, and we used the fact that $H_\text{lin} = m\dot{x}_\text{CoM}$. 
 
In order to model the step in a model predictive framework, we can assume to know the instant in which the swing foot will touch the ground.
In fact, the considered model does not contain any information about the posture of the robot and therefore it is not possible to define a ``transition function'', e.g. the distance from the foot to the ground is generally used to predict when the step is going to take place. 
In other words, the controller is aware that the impact will take place in a precise instant in the future, but it does not know whether the quantities involved in the model will affect this instant or not.
The most viable choice is then to consider this instant as constant and known in advance, equal to $t_{impact}$, i.e. to the time needed to perform a step.

For the same reason, the position that the swing foot will have after the step is also assumed to be known and constant. 
This takes particular importance since ${}^{\text{CoM}}X_{s}$ (where $s$ refers to the swing foot) directly depends on this position.
Summarizing, the peculiar characteristics of the step, i.e. the duration and the target position, are assumed to be known and constant. 

\subsection{The angular momentum}\label{sec:ang_mom}
We focus now on the matrix ${}^{\text{CoM}}X_{i}$ (here $i$ is a placeholder for subscript $l$ or $r$), whose structure depends on the choice of the frame to describe the robot momentum. 
The matrix has the following form:
\begin{equation}
{}^{\text{CoM}}X_{i} = \begin{bmatrix}
								{1}_3 & 0_{3\times 3}\\
								(x_{i}-x_{\text{CoM}})^{\wedge} & {1}_3
						\end{bmatrix}.
\end{equation} 
The derivative of the angular momentum $\dot{H}_{\text{ang}}$ may thus be written as
$
\dot{H}_\text{ang} = \sum_{i}^{l,r} (x_{i}-x_{\text{CoM}})^\wedge \: \mathbbm{f}_i+\tau_i .
$
The product between $x_\text{CoM}$ and $\mathbbm{f}_i$ causes the whole formulation to be non-convex which makes the resolution dramatically more complex.
In literature this problem is faced by minimizing an upper-bound of the angular momentum \cite{daiplanning}, or by approximating it through quadratic constraints \cite{ponton2016convex}. 
In our application, angular momentum is mainly needed to bound the usage of contact wrenches, rather than be precisely controlled to zero. Thus, we can accept a more coarse approximation, relying on its Taylor expansion (around the last available values of $\mathbbm{f}_i$ and $x_{\text{CoM}}$) truncated to the first order, namely: 
\begin{IEEEeqnarray*}{RCL}
\IEEEyesnumber \phantomsection
\dot{H}_\text{ang} &\approx& \sum_{i}^{\{l,r\}} \dot{H}_\text{ang}
 + \frac{\partial \dot{H}_\text{ang}}{\partial \mathbbm{f}_i} \left(\mathbbm{f}_i - \mathbbm{f}_i^0\right)+ \frac{\partial \dot{H}_\text{ang}}{\partial x_\text{CoM}} \left(x_\text{CoM} - x_{\text{CoM}}^0\right) \nonumber\\
\dot{H}_\text{ang} &\approx& \sum_{i}^{\{l,r\}} \tau_i + \left(x_{i}-x_{\text{CoM}}^0\right)^\wedge \: \mathbbm{f}_i^0+  \left(x_{i}-x_{\text{CoM}}^0\right)^\wedge \: \left(\mathbbm{f}_i-\mathbbm{f}_i^0\right) +  \nonumber \\
&+& \left(\mathbbm{f}_i^0\right)^\wedge \: \left(x_{\text{CoM}}-x_{\text{CoM}}^0\right).  \label{eq:ang_mom_last} \nonumber\\
\Rightarrow \dot{H}_\text{ang} &\approx& \sum_{i}^{\{l,r\}} \tau_i + \left(x_{i}-x_{\text{CoM}}^0\right)^\wedge \: \mathbbm{f}_i + \left(\mathbbm{f}_i^0\right)^\wedge \: \left(x_{\text{CoM}}-x_{\text{CoM}}^0\right) \IEEEyesnumber
\end{IEEEeqnarray*}
where we exploited the anticommutative property of the cross product, i.e.  $A^\wedge B = -B^\wedge  A$. The superscript ${}^0$ refers to the point at which the Taylor expansion is computed, in our case from the last feedback obtained from the robot. The approximation introduced with the truncation to the first order is  $o\left(\left(\mathbbm{f}_i-\mathbbm{f}_i^0\right)\left(x_{\text{CoM}}-x_{\text{CoM}}^0\right) \right)$. 
If we consider a controller horizon $t_f$ not too wide, the term related to the position of the CoM is expected to be small enough, making this linear approximation effective.

Finally, by defining the state variable $\gamma$ and the control input variable $f$ as
$\gamma := \left[\begin{smallmatrix}
x_\text{CoM}^\top &
\dot{x}_\text{CoM}^\top & 
H_\text{ang}^\top
\end{smallmatrix}\right]^\top,~ f := \left[\begin{smallmatrix}
f_l^\top &
f_r^\top
\end{smallmatrix}\right]^\top
$
we can rewrite Eq. \eqref{c6eq:model_2f} as:
\begin{equation}\label{c6eq:model}
\dot{\gamma} = \tilde{E}v_\gamma \gamma + \tilde{F}_\gamma f + \tilde{G}_\gamma + \tilde{S}_\gamma^0,
\end{equation}
\begin{IEEEeqnarray*}{RCL}
\tilde{E}v_\gamma &=& \begin{bmatrix}
0_{3\times3} & {1}_3 & 0_{3\times3}\\
0_{3\times3} & 0_{3\times3}  & 0_{3\times3}\\
\left(\mathbbm{f}_l^0+\mathbbm{f}_r^0\right)^\wedge & 0_{3\times3}  & 0_{3\times3}\\
\end{bmatrix}, \tilde{F}_\gamma {=} \begin{bmatrix}
	                        0_{3\times 3} &0_{3\times 3} &0_{3\times 3} &0_{3\times 3} \\
				           m^{-1}1_3 & 0_{3\times 3} & m^{-1}1_3 & 0_{3\times 3}\\
				           \left(x_{l}-x_{\text{CoM}}^0\right)^\wedge & 1_3 & \left(x_{r}-x_{\text{CoM}}^0\right)^\wedge & 1_3					           		
			        \end{bmatrix},\\
\tilde{G}_\gamma &=& \begin{bmatrix}
0_3\\
-\bar{g}
\end{bmatrix},\quad \tilde{S}^0_\gamma = \begin{bmatrix}
							  0_6\\
							  -\left(\mathbbm{f}_l^0+\mathbbm{f}_r^0\right)^\wedge x_{\text{CoM}}^0
                       \end{bmatrix}.
\end{IEEEeqnarray*}

%% file: tex/constraints.tex
\subsection{Constraints definition}\label{c6sec:constraints}

The constraints to be enforced are mainly related to the feasibility of the exerted wrenches, e.g. center of pressure and friction cone among the most common constraints related to unilateral contacts.
Furthermore the wrench acting on the swinging foot should be null for every instant before the impact, i.e. $t_{impact}$.

We start by examining the constraints acting on the stance foot (here assumed to be the left):
\begin{equation}\label{c6eq:wrench_constr}
A_{cl}f_l \leq b_{cl} \quad \forall t: t \leq t_f.
\end{equation}
Eq. \eqref{c6eq:wrench_constr} encompasses all the considered inequalities constraints, namely: 
\begin{inparaenum}[i)]
    \item friction cone with linear approximation,
    \item center of pressure,
    \item positivity of the normal component of the contact force.
\end{inparaenum}

The constraints on the swing foot instead, catches the hybrid nature of the system. In particular they represent the fact that the wrench must be null before the impact and feasible after it. Formally:
\begin{equation}\label{c6eq:fr_cont_costr}
\begin{cases}
	f_{r} = 0_6 &\forall t:\; t < t_{impact}\\
	A_{cr}f_r \leq b_{cr}  &\forall t:\; t_{impact} \leq t \leq t_f
\end{cases}
\end{equation}
where $A_{cr}$ and $b_{cr}$ describe the same feasibility constraints as $A_{cl}$ and $b_{cl}$.
The above equation assumes that $t_{impact} \leq t_f$. If the impact does not occur inside the control horizon, then the wrench exerted by the right foot is forced to be null throughout the whole horizon.

We also included an additional constraint to enforce that balancing is kept after the step. In particular, after this instant, we can constraint the instantaneous capture point to be inside the convex hull of the two feet, which can be predicted by knowing the future position of the right foot.
Thus, we can define a set of linear inequalities such that if
$
A_{ch}x_{icp} \leq b_{ch} \quad \forall t: t \geq t_{impact}
$
is satisfied, than $x_{icp}$, i.e. the instantaneous capture point location, is in the convex hull.
By imposing this constraint, we are forcing the instantaneous capture point to be in a \emph{capturable} state after the step is performed. In fact, the convex hull represents the region in which it is possible to stabilize the instantaneous capture point dynamics without taking additional steps.

%% file: tex/cost_fcn.tex
\subsection{Cost function definition}\label{c6sec:cost}
Following the optimal control formulation of Eq. \eqref{c6eq:ocdef} we are left to define the cost function applied within the MPC controller, called ${\Gamma}$.
Note that different terms of the cost function act only after the step is taken. Formally, it has the following expression:
\begin{IEEEeqnarray}{rl}
\IEEEyesnumber \phantomsection \label{c6eq:cost}
&{\Gamma} = \frac{1}{2} \left(  \int_{0}^{t_f} \norm{ \gamma(\tau) - \gamma^d(\tau)}^2_{K_\gamma}  \dif\tau + \int_{\bar{t}_{imp}}^{t_f} \norm{\gamma(\tau) - \gamma^d(\tau)}^2_{K_\gamma^{imp}}  \dif\tau + \right. \IEEEyesnumber \label{c6eq:cost_gamma2}\\
 &+\int_{0}^{t_f}  \norm{f(\tau)}^2_{K_f} \dif\tau +\int_{\bar{t}_{imp}}^{t_f}  \norm{x_{icp}(\tau) - x_{icp}^d(\tau)}^2_{K_{icp}} \dif\tau + \norm{\gamma(t_f) - \gamma^d(t_f)}^2_{K_\gamma^{imp}}\Bigg) \nonumber
\end{IEEEeqnarray}
where the superscript $d$ indicates the reference values.
$K_{(\cdot)}^{(\cdot)}$ is a real positive semi-definite matrix of gains with suitable dimensions, accounting also for unit of measurement mismatches inside ${\Gamma}$. $\bar{t}_{imp}$ is the minimum between $t_{impact}$ and $t_f$ while, for the sake of simplicity, the initial time instant is set to zero.
	Thus, it is possible to vary the cost applied to the state $\gamma$ after the impact. This is done through the use of the matrix $K_\gamma^{imp}$ which, after the impact, adds up to $K_\gamma$. 
Finally, a terminal cost term is inserted using the same matrix $K_\gamma^{imp}$, to model the finiteness of the control horizon.

Consider now the final objective of having the center of mass over the centroid of the support polygon once the step is made. We decide to weight only the $z-$component of the CoM error throughout the whole horizon, while weighting the traverse components (i.e. $x$ and $y$) only in the terminal cost (last term of Eq.\eqref{c6eq:cost_gamma2}) and after the step is made (second part of Eq.\eqref{c6eq:cost_gamma2}). 
The same rationale could be applied to the reference for the instantaneous capture point. During the step, its dynamics is exponentially diverging and, as a consequence, this contribution takes a role only after $t_{impact}$, where its distance from the centroid of the convex hull is weighted.

Finally, the requested reaction forces are weighted too, mostly for avoiding impulsive responses which are dangerous for the mechanical structure of the robot, without any warranty that they will be actually attained through the underneath controller. Intuitively, the tracking error for the desired wrenches will be lower with a smooth and limited reference.

%% file: tex/simulation.tex
\section{Simulation Results} \label{sec:simulation}
\begin{figure}[t]
	\centering
	\subfloat[]{\includegraphics[width=.45\columnwidth]{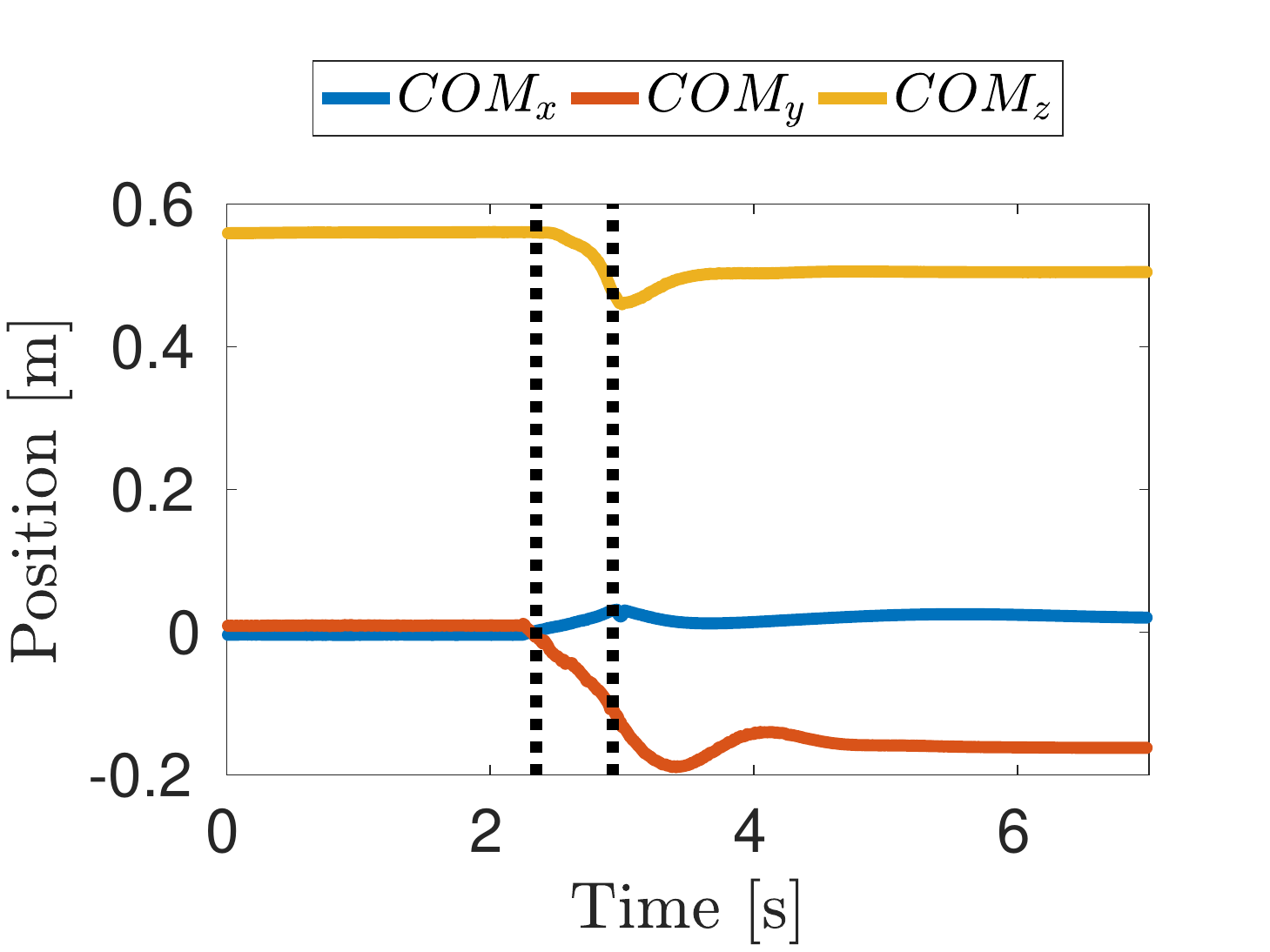}}
	\subfloat[]{\includegraphics[width=.45\columnwidth]{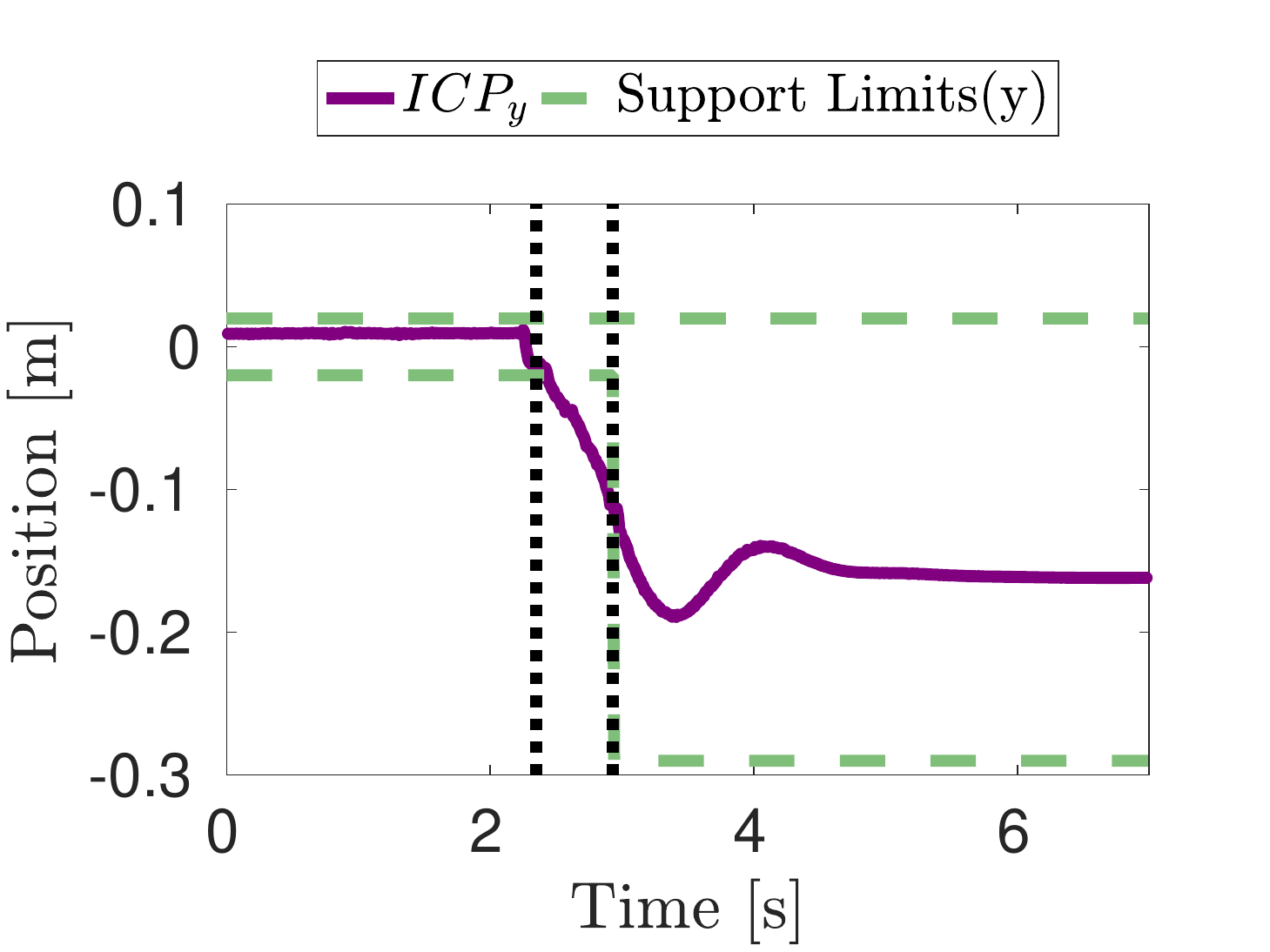}}
	\caption{(a) represents the center of mass evolution while performing the step. The first dotted line denotes the time of the push which occurred at $t = 2.4\mathrm{s}$. The second one at $t = 3.0 \mathrm{s}$ indicates when the right foot hits the ground. (b) presents the evolution of the $y$-component of the instantaneous capture point. The horizontal green dotted lines represent the approximation of the convex hull.}\label{c6fig:com}
\end{figure}

\begin{figure*}[t]
	\centering
	\includegraphics[width=.45\columnwidth]{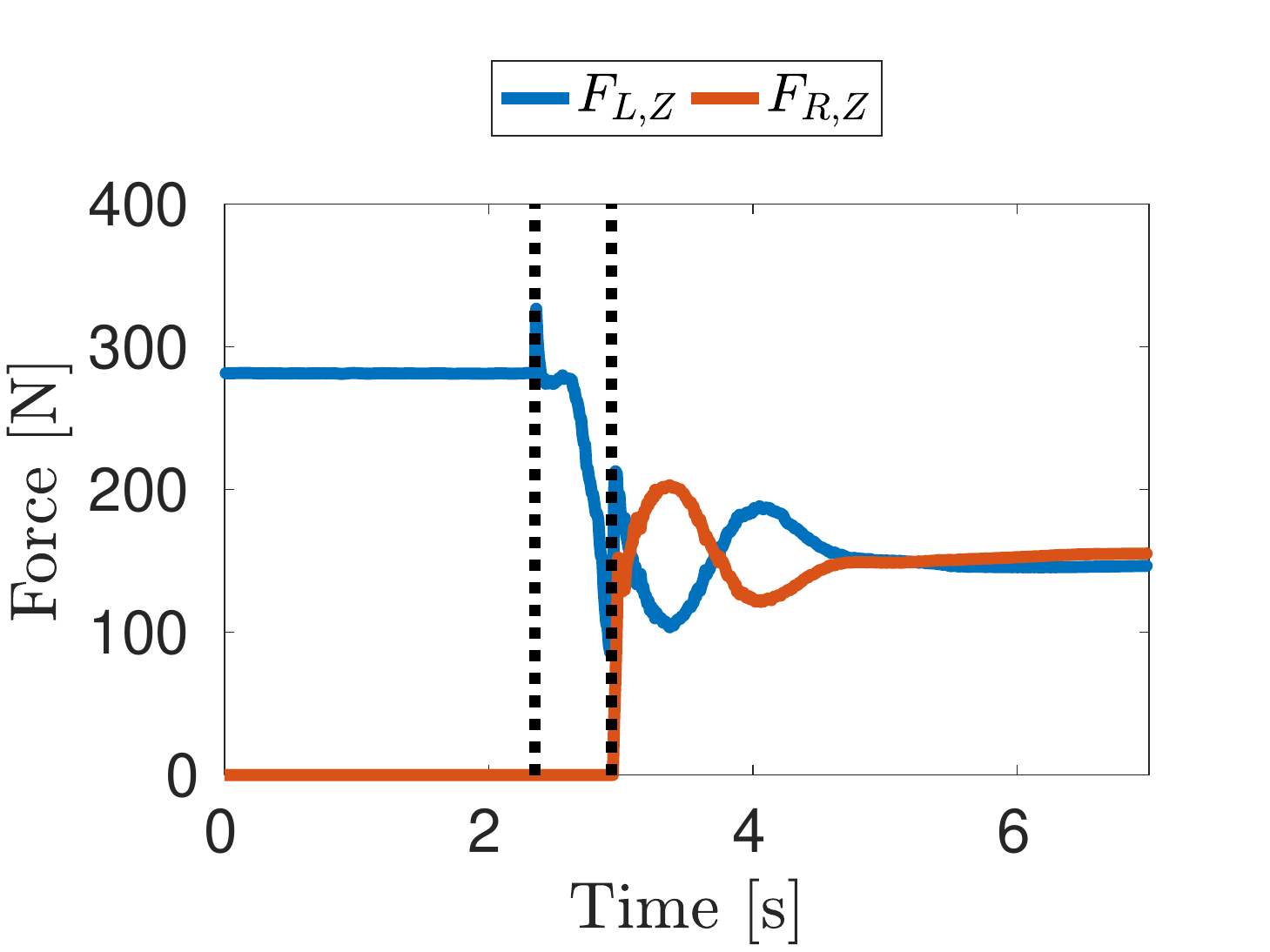}
	\includegraphics[width=.45\columnwidth]{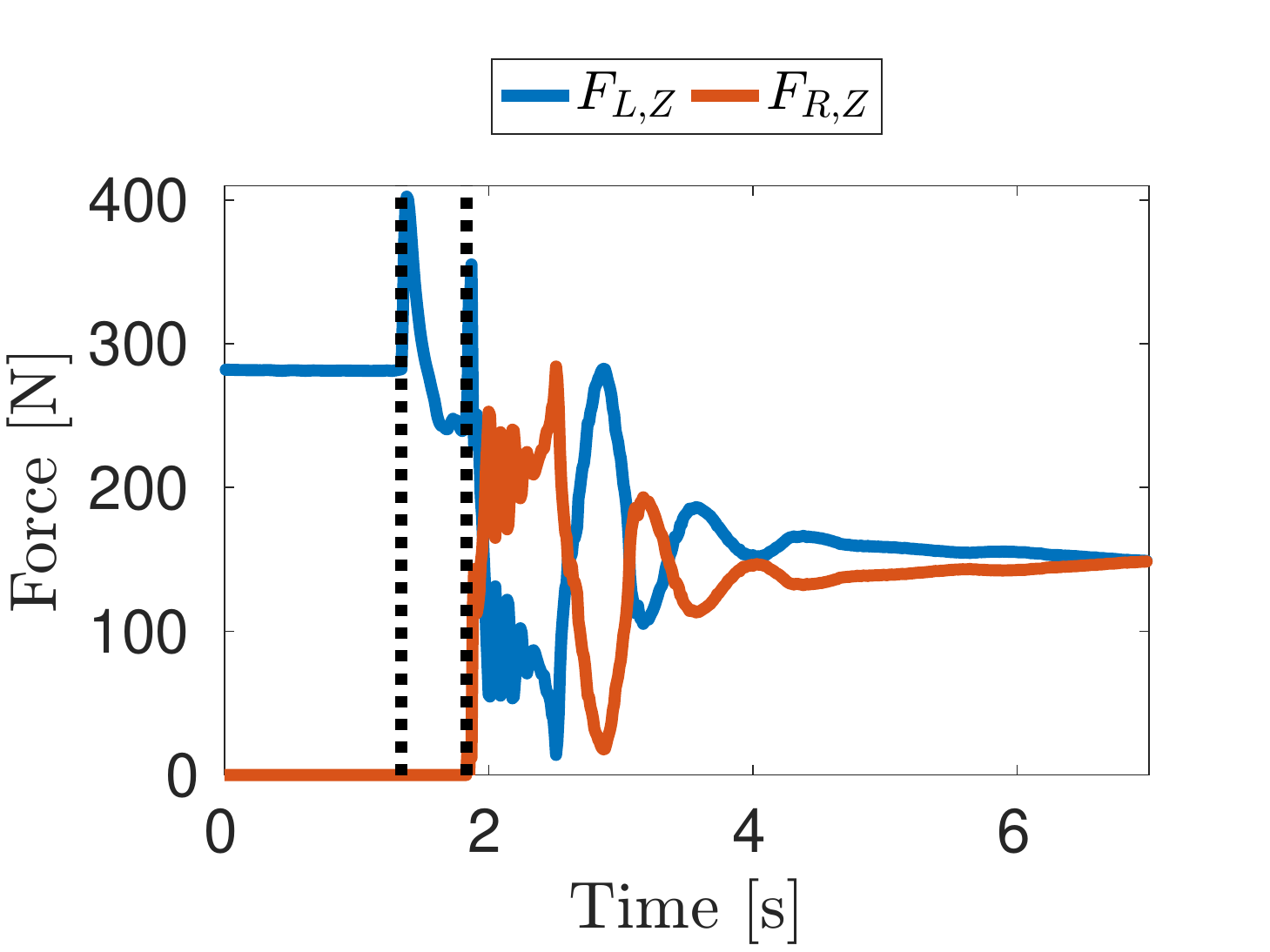}
	
	\caption{Representation of the $z$-component of the demanded forces.}
	\label{c6fig:force}
\end{figure*}

The proposed MPC approach has been tested on an iCub model in the Gazebo simulator \cite{Koenig04}, employing YARP Plugins \cite{YarpGazebo2014}. 

The iCub humanoid robot \cite{metta2005robotcub} possesses 53 actuated joints, but nearly half of them are concentrated on the eyes and the hands. For this reason only 23 degrees of freedom (DoF) are torque controlled and used for balancing purposes.

The proposed controller has been implemented in the MATLAB\textsuperscript{\textregistered}/Simulink\textsuperscript{\textregistered}  environment, using the WBToolbox library\cite{RomanoWBI17}.
The experimental scenario consists of the robot balancing on its left foot. We have chosen this simple scenario to test the performance of the presented strategy with a single contact activation.

When a strong push occurs the step recovery strategy starts. The time step $\mathrm{d}t$ is set to 10ms, while the controller horizon $N$ is chosen to be 15. It has been observed that for smaller control horizons, the robot is not able to sustain the disturbance. 
Figure \ref{c6fig:com} depicts the result of the push experiment performed adopting the MPC approach. 
The robot was able to fully recover after the push, undergoing just a slight oscillation after the impact of the swing foot, whose occurrence is highlighted by the second dashed line. The capability of recovering by performing the step can be noticed also from Figure \ref{c6fig:com}, where the behavior of the $y$-component of the Instantaneous Capture Point is depicted (the $x$ component has undergone just a slight perturbation). In particular the Capture Point, after exiting the support polygon (corresponding to the left foot), is fully contained in the new convex hull once the step is performed.

Finally, it is worth considering the desired wrenches. 
Figures \ref{c6fig:force} compare the desired vertical force obtained using the presented control strategy and the one proposed in \cite{Dafarra2016} (on the right).
While oscillating in the first stage of the double support, the steady state is reached much faster than the previous approach.

%% file: tex/conclusions.tex
\section{Conclusions and Future Work} \label{sec:conclusion}
The presented approach adopts, as model, the dynamics of the center of mass with only a small approximation necessary to remove non-convexities. 
It also considers the discontinuous contribution of the swing foot through the adoption of time varying constraints. 
With respect to \cite{Dafarra2016}, this approach does not need the definition of a trajectory for the CoM, which indeed is part of the optimization process. 
As a consequence, the robustness of this technique relies more in the intrinsic knowledge of the system rather than in the expertise of the designer.

This approach has been meant not to be a simple planner, but instead to provide reaction to external disturbances in real-time. 
However, at the present time it takes almost $0.1$ seconds to get a solution on a machine running Ubuntu 14.04 on a quad-core Intel\textsuperscript{\textregistered} Core i5@2.30GHz with 16GB of RAM, by using as solver the MATLAB\textsuperscript{\textregistered} interface of MOSEK\textsuperscript{\textregistered}. As a future work, this strategy will be implemented on an optimized C++ code and applied on the real platform. 

Finally, as an additional improvement we plan to supply the controller with more informations about the kinematics of the leg, thereby modeling the step movement directly as part of the problem formulation and relaxing the hypothesis of knowing a priori the time of the impact.